\pdfoutput=1

\documentclass[11pt]{article}

\usepackage{acl}

\usepackage{times}
\usepackage{latexsym}

\usepackage[T1]{fontenc}

\usepackage[utf8]{inputenc}

\usepackage{microtype}

\usepackage{inconsolata}

\usepackage{graphicx}
\usepackage{booktabs}
\usepackage{url}

\usepackage{CJKutf8}
\usepackage{makecell}
\title{FollowEval: A Multi-Dimensional Benchmark for Assessing the Instruction-Following Capability of Large Language Models}

\author{
Yimin Jing$^{1}$\thanks{~Equal contribution}\hspace{0.5em},
Renren Jin$^{2}$\footnotemark[1]\hspace{0.5em}, 
Jiahao Hu$^{1}$,
Huishi Qiu$^{1}$\\
\textbf{Xiaohua Wang}$^{1}$,
\textbf{Peng Wang}$^{1}$,
\textbf{Deyi Xiong}$^{2}$\\
Lenovo Research AI Lab$^{1}$, Tianjin University$^{2}$\\
\texttt{\{jingym3,hujh9,qiuhs1,wangxh40,wangpeng31\}@lenovo.com}\\
\texttt{\{rrjin,dyxiong\}@tju.edu.cn}
}

\begin{document}
\maketitle

\begin{abstract}
The effective assessment of the instruction-following ability of large language models (LLMs) is of paramount importance. A model that cannot adhere to human instructions might be not able to provide reliable and helpful responses. In pursuit of this goal, various benchmarks have been constructed to evaluate the instruction-following capacity of these models. However, these benchmarks are limited to a single language and are constructed using automated approaches, which restricts their applicability and the quality of the test examples they contain. To bridge this gap, we introduce the FollowEval benchmark in this paper. This benchmark is composed of instances in both English and Chinese, and all test examples are crafted by human experts. Furthermore, the FollowEval benchmark is designed to assess LLMs across five critical dimensions of instruction following: string manipulation, commonsense reasoning, logical reasoning, spatial reasoning, and response constraints. To enhance the complexity and present a sufficient challenge, each test example is designed to evaluate more than one dimension. We have evaluated various LLMs using the FollowEval benchmark and found that their performance significantly lags behind that of humans. This highlights the considerable room for improvement in the instruction-following ability of these models.
\end{abstract}

\section{Introduction}

Large language models (LLMs) have gained substantial attention due to their remarkable performance across a diverse range of tasks \citep{DBLP:journals/corr/abs-2303-12712,DBLP:journals/corr/abs-2304-14106,DBLP:journals/corr/abs-2301-08745,DBLP:journals/corr/abs-2303-04048,DBLP:journals/corr/abs-2305-14835}. A fundamental technique that endows LLMs with such exceptional capabilities is instruction tuning. This technique aligns the responses of LLMs with human values, thereby enabling these models to adhere to instructions \citep{DBLP:conf/nips/Ouyang0JAWMZASR22,DBLP:journals/corr/abs-2204-05862,DBLP:journals/corr/abs-2304-03277,DBLP:journals/corr/abs-2305-11206,DBLP:journals/corr/abs-2305-18290}. By enabling LLMs to follow natural language instructions, this technique facilitates more natural human-AI interaction and enables the LLMs to reliably accomplish the tasks as specified in human instructions. Consequently, the ability of LLMs to follow instructions significantly enhances their reliability and utility in real-world applications. Therefore, it becomes imperative to assess the instruction-following proficiency of LLMs before their deployment.

The demand for a comprehensive evaluation of the instruction-following ability of LLMs has stimulated the construction of various benchmarks aimed at assessing the instruction-following capacity of LLMs in recent years \citep{DBLP:journals/corr/abs-2212-10466,DBLP:journals/corr/abs-2307-08689,DBLP:journals/corr/abs-2309-09150,DBLP:journals/corr/abs-2311-04235,DBLP:journals/corr/abs-2310-14542,DBLP:journals/corr/abs-2310-20410}. However, these benchmarks exclusively focus on either English or Chinese and are constructed based on automated approaches, which restricts their practical applications and the quality of the test instances in their benchmarks.

In response to this demand and the aforementioned limitations, we present the FollowEval benchmark in this paper, which is specifically designed to evaluate the instruction-following ability of LLMs. Unlike existing benchmarks, the FollowEval benchmark covers both English and Chinese, and all test instances within it are manually curated. This expands its scope of usage and ensures the quality of this benchmark. As shown in Table~\ref{test_examples_1}, there are three notable features in the FollowEval benchmark: (1) It can evaluate LLMs across diverse dimensions within instruction-following, such as string manipulation and commonsense reasoning, referred to as essential elements in our paper. (2) Each test instance in the FollowEval benchmark incorporates more than one essential element, and the response of the LLMs is deemed incorrect unless all essential elements are adequately addressed, thereby increasing the difficulty of the FollowEval benchmark. (3) Each test instance is associated with a manually designed rule implemented by regex to verify the correctness of the LLMs' responses, which facilitates lightweight evaluation and ensures the reliability of the evaluation.

Our contributions can be summarized as follows:
\begin{itemize}
    \item We have curated the FollowEval benchmark to evaluate the instruction-following ability across diverse dimensions of LLMs. This includes 200 manually written test instances that cover both English and Chinese. Furthermore, we provide the corresponding manually designed regex to facilitate the evaluation.
    \item We have conducted evaluations of various LLMs on the FollowEval benchmark. The experimental results reveal a substantial gap between the instruction-following capacity of LLMs and humans, highlighting areas for improvement.
\end{itemize}

\begin{CJK*}{UTF8}{gbsn}
\begin{table}[!ht]
\small
\centering
\begin{tabular}{p{3.5cm}l}
\toprule
\textbf{Instruction} & \textbf{Evaluation Regex} \\
\midrule
\multicolumn{1}{m{3.5cm}}{请给我输出5个K，不要生成其他任何额外字符，且字母直接相连，没有任何分隔符\ (\textit{Please generate five consecutive `K's without adding any extra characters or separators})} & \makecell[l]{if answer == "KKKKK":\\\ \ \ \ \ \ return 1\\else:\\ \ \ \ \ \ \ return 0} \\
\midrule
\multicolumn{1}{m{3.5cm}}{请生成一句话，以英文字母开头，以``我''字结尾\ (\textit{Please generate a sentence that starts with an English letter and ends with the word ``我''})} & \makecell[l]{if re.search(\\r"\^~[a-zA-Z].*我\$", \\answer\\):\\ \ \ \ \ \ \ return 1\\else:\\ \ \ \ \ \ \ return 0} \\
\midrule
\multicolumn{1}{m{3.5cm}}{若a > b，b > c，请问a > c吗，直接回答正确还是错误，不要输出分析过程和额外的其他字符\ (\textit{If `a' is greater than `b', and `b' is greater than `c', then is it correct to say that `a' is greater than `c'? Please respond with `correct' or `incorrect' directly, without including any analytical process or extra characters})} & \makecell[l]{if answer == "正确":\\ \ \ \ \ \ \ return 1\\else:\\ \ \ \ \ \ \ return 0} \\
\midrule
\multicolumn{1}{m{3.5cm}}{请将句子``天空中的太阳非常明亮''中的所有``的''字去除，并重新输出\ (\textit{Please eliminate all instances of ``的'' from the sentence ``天空中的太阳非常明亮'' and then reproduce it})} & \makecell[l]{if "天空中太阳非常\\明亮" in answer:\\ \ \ \ \ \ \ return 1\\else:\\ \ \ \ \ \ \ return 0} \\
\midrule
\multicolumn{1}{m{3.5cm}}{``我想要在绿色的草原上，顶着红色的太阳漫步''，请将句中表示颜色的两个汉字交换并重新输出\ (\textit{Please swap the two Chinese characters that denote color in the sentence and then reproduce it. The sentence is: ``我想要在绿色的草原上，顶着红色的太阳漫步''})} & \makecell[l]{if "我想要在红色的\\草原上，顶着绿色\\的太阳漫步" \\in answer:\\ \ \ \ \ \ \ return 1\\else:\\ \ \ \ \ \ \ return 0} \\
\midrule
\multicolumn{1}{m{3.5cm}}{强的反义词和大的反义词连在一起是什么词\ (\textit{What is the word formed by combining the antonyms of ``强'' and ``大''?)}} & \makecell[l]{if "弱小" in answer:\\ \ \ \ \ \ \ return 1\\else:\\ \ \ \ \ \ \ return 0} \\
\bottomrule
\end{tabular}
\caption{Examples with the associated instructions and corresponding evaluation regex from FollowEval benchmark.}
\label{test_examples_1}
\end{table}
\end{CJK*}

\section{Related Work}
Given the remarkable proficiency of LLMs in both understanding and generating language, there has been a significant increase in research efforts aimed at curating benchmarks to assess the capacity of LLMs to follow instructions. The primary focus of \citet{DBLP:journals/corr/abs-2212-10466} is on the knowledge-intensive constraints, and the corresponding benchmarks are constructed automatically, leveraging the resources of WordNet \citep{DBLP:journals/cacm/Miller95} and Wikidata \citep{DBLP:journals/cacm/VrandecicK14}. Subsequently, \citet{DBLP:journals/corr/abs-2307-08689} expand the types of constraints through a grammar-based framework, which facilitates the automatic formulation of instructions with compositional constraints across diverse generative levels. In contrast to previous works that construct benchmarks by leveraging existing resources or rule-based frameworks, \citet{DBLP:journals/corr/abs-2309-09150} initially collect instructions from practical scenarios. They then prompt the LLMs to diversify these instructions and augment their complexity. Concurrent with our work, \citet{DBLP:journals/corr/abs-2311-04235,DBLP:journals/corr/abs-2310-14542,DBLP:journals/corr/abs-2310-20410} construct benchmarks for evaluating the instruction-following capabilities of LLMs. However, \citet{DBLP:journals/corr/abs-2311-04235} primarily focus on measuring the instruction-following ability of LLMs under adversarial inputs. \citet{DBLP:journals/corr/abs-2310-14542} introduce the numerical planning benchmark to assess the capacity of large LLMs to generate texts that satisfy numerical constraints, such as word count and syllable count. Furthermore, to the best of our knowledge, these existing instruction-following benchmarks are restricted to English or Chinese and are constructed with the assistance of automated approaches. In contrast, the examples in the FollowEval benchmark are human-authored and span both English and Chinese.

\section{FollowEval}

\begin{CJK*}{UTF8}{gbsn}
\begin{table*}[!ht]
\small
\centering
\begin{tabular}{p{9cm}p{6cm}}
\toprule
\textbf{Instruction} & \textbf{Essential Elements} \\
\midrule
请用``雨伞''和``跳舞''这两个词汇中的第一个词汇造句\ (\textit{Please make a sentence using the first of these two words: ``雨伞'' and ``跳舞''}) & Character Position Identification, Commonsense Reasoning\\
\midrule
\multicolumn{1}{m{9cm}}{使用``乐不思蜀''这个成语造句，且造出的句子有且只有15个字符\ (\textit{Please create a sentence with only 15 characters using the idiom ·`乐不思蜀'‘})} & Commonsense Reasoning, Length Constraints \\
\midrule
\multicolumn{1}{m{9cm}}{请输出1个a，两个B，三个C，四个d顺序组成的字符串，注意区分大小写\ (\textit{Please output a string that consists of 1 `a', 2 `B's, 3 `C's, and 4 `d's in that order, keeping in mind that case sensitivity matters})} & Character Counting, Formality Constraints \\
\midrule
\multicolumn{1}{m{9cm}}{请输出元素周期表前5个元素的英文简写，每个元素用空格隔开\ (\textit{Please provide the English abbreviations for the first five elements of the periodic table, with each element separated by a space})} & \multicolumn{1}{m{6cm}}{Commonsense Reasoning, Formality Constraints, Character Position Identification} \\
\midrule
\multicolumn{1}{m{9cm}}{把"甲"倒过来是什么字？\ (\textit{What word does ``甲'' become when it’s turned upside down?})} & Commonsense Reasoning, Character Rotation \\
\midrule
\multicolumn{1}{m{9cm}}{假设你现在有4张卡片A、B、C、D，A卡片上写着happy，B卡片写着want，C卡片写着very，D卡片写着basketball。请使用A卡片和B卡片上的单词造一个句子\ (\textit{Assume you have four cards labeled A, B, C, and D. `happy' is written on card A, `want' on card B, `very' on card C, and `basketball' on card D. Please construct a sentence using the words on cards A and B})} & Commonsense Reasoning, Spatial Reasoning \\
\bottomrule
\end{tabular}
\caption{Examples with the associated instructions and corresponding essential elements from FollowEval benchmark.}
\label{test_examples_2}
\end{table*}
\end{CJK*}

\subsection{Desgin Principle}

The construction of an instruction-following benchmark for LLMs poses several significant challenges:

\paragraph{Benchmark Curation} The benchmark must comprise a comprehensive, representative set of instructions that align with common real-world applications, while also sufficiently challenging the capabilities of current LLMs. Careful curation of the benchmark examples is necessary to satisfy these dual objectives.

\paragraph{Automatic Evaluation} The automatic evaluation of natural language generation remains an open challenge due to the complexity and diversity of natural language. While the use of LLMs to evaluate the quality of responses they produce shows promising avenues, this approach has major limitations. Specifically, LLMs tend to: (1) Score responses at certain positions higher when multiple responses are provided simultaneously for evaluation \citep{DBLP:journals/corr/abs-2305-17926,DBLP:journals/corr/abs-2306-05685}; (2) Prefer responses that are long, verbose, and contain many unique tokens \citep{DBLP:journals/corr/abs-2306-04751,DBLP:journals/corr/abs-2306-05685}; (3) Favor responses that they have generated themselves \citep{DBLP:journals/corr/abs-2303-16634,DBLP:journals/corr/abs-2306-05685}.

\paragraph{Environmental and Financial Costs} LLM-based evaluations can lead to non-negligible carbon emissions due the significant computational cost during LLM inference. Additionally, if proprietary models that provide services through APIs, such as ChatGPT and GPT-4, are used as evaluators, there are associated API costs.

To mitigate these challenges, we have predefined five essential elements that are closely related to the practical usage scenarios that the FollowEval benchmark aims to cover. Precisely responding to each of these five elements is key for LLMs to generate fully correct responses. Moreover, the assessment of the LLMs’ responses to test examples containing these five essential elements can be easily accomplished through handcrafted rules, thereby facilitating a reliable, convenient, and cost-effective evaluation. The five essential elements we have defined are as follows:

\paragraph{String Manipulation} String manipulation refers to the technique of analyzing and managing strings, Given that LLMs operate with text data, which is composed of strings, string manipulation enables LLMs to effectively analyze, modify, and synthesize textual information. Consequently, it is a fundamental competency for LLMs. For the FollowEval benchmark, the string manipulation elements include several sub-elements: position identification, character insertion, character deletion, and character replacement.

\paragraph{Commonsense Reasoning} Commonsense reasoning enhances the capacities of LLMs beyond the narrow processing of literal language to broadly capable, assistive AI systems that demonstrate more human-like comprehension and reasoning. Therefore, it is a key component to enable LLMs to understand and generate text in a manner that is more aligned with human cognition. For test examples with the commonsense reasoning element in the FollowEval benchmark, the LLMs must possess the relevant commonsense knowledge to produce appropriate responses.

\paragraph{Logical Reasoning} Logical reasoning refers to the ability to comprehend statements, interpret them through logical analysis and theoretical establishment, which enhances the consistency and reliability of the responses generated by LLMs \citep{DBLP:journals/corr/abs-2310-09158}. For the FollowEval benchmark, mathematical computation and character counting are incorporated as sub-elements of logical reasoning.

\paragraph{Spatial Reasoning} Spatial reasoning refers to the capacity to conceptualize objects in both two and three dimensions and draw conclusions from the given information. Incorporating tests of spatial reasoning into benchmarks for LLMs could provide a more comprehensive evaluation. For the FollowEval benchmark, there are two sub-elements in the spatial reasoning element: spatial transformations and character rotation.

\paragraph{Response Constraints} The imposition of well-designed response constraints in the instructions provided to LLMs can guide the LLMs to produce more accurate, relevant, and helpful responses that meet user requirements. Given that setting response constraints in the instructions is a common practice when interacting with LLMs, testing how well LLMs adhere to specified response constraints is a significant aspect of evaluating LLMs' instruction-following capabilities. For the FollowEval benchmark, the response constraints element includes three sub-elements: length constraints, formality constraints, and character constraints.

In real-world scenarios, instructions typically encompass multiple essential elements that must be concurrently addressed by LLMs. Consequently, to conduct a comprehensive evaluation of the instruction-following capabilities of LLMs across various dimensions, every test instance in the FollowEval benchmark is designed to incorporate more than one essential element. This intentional design strategy serves to augment the complexity and challenge of the FollowEval benchmark. Table~\ref{test_examples_2} presents some test examples from the FollowEval benchmark, along with the corresponding essential elements for each example. Moreover, considering that Chinese represents the language with the largest number of native speakers globally, and English is the most widely spoken language across the world, our current research within the FollowEval benchmark is focused primarily on these two languages. Incorporating more languages into the FollowEval benchmark is left for future work.

\subsection{Dataset Construction}
To ensure the quality of the FollowEval benchmark, we have devised a three-step curation process: instruction drafting, instruction verification, and regular expression design. This systematic approach ensures a comprehensive and rigorous evaluation framework. The curation process is conducted entirely by human experts, with six individuals drafting the initial instructions, two separate experts verifying the drafted instructions, and another two specialists designing the corresponding regular expressions for each instruction.


\begin{table*}[]
\small
\centering
\resizebox{\textwidth}{!}{
\begin{tabular}{@{}lccccccc@{}}
\toprule
Model & \multicolumn{3}{c}{Chinese Samples} & \multicolumn{3}{c}{English Samples} & \multicolumn{1}{l}{} \\ \midrule
 & ACC-1 & ACC-2 & ACC-3 & ACC-1 & ACC-2 & ACC-3 & AVG \\
Human &  &  &  &  &  &  &  1.000 \\
GPT-4 & 0.83 & 0.82 & 0.78 & 0.74 & 0.73 & 0.75 & 0.775 \\
GPT-3.5-Turbo & 0.64 & 0.64 & 0.64 & 0.66 & 0.62 & 0.70 & 0.650 \\
Qwen-14B-Chat & 0.62 & 0.62 & 0.61 & 0.44 & 0.44 & 0.42 & 0.525 \\
Qwen-7B-Chat & 0.54 & 0.57 & 0.53 & 0.34 & 0.36 & 0.36 & 0.448 \\
Baichuan-2-13B-Chat & 0.42 & 0.43 & 0.45 & 0.38 & 0.39 & 0.39 & 0.408 \\
Baichuan-13B-Chat & 0.40 & 0.47 & 0.41 & 0.27 & 0.30 & 0.29 & 0.357 \\
ChatGLM3-6B & 0.38 & 0.39 & 0.36 & 0.35 & 0.32 & 0.32 & 0.353 \\
InternLM-7B-Chat & 0.32 & 0.31 & 0.31 & 0.36 & 0.41 & 0.34 & 0.345 \\
Baichuan-2-7B-Chat & 0.35 & 0.38 & 0.34 & 0.30 & 0.32 & 0.33 & 0.337 \\
ChatGLM2-6B & 0.34 & 0.32 & 0.30 & 0.27 & 0.24 & 0.26 & 0.288 \\
LLaMA-2-13B-Chat & 0.25 & 0.26 & 0.27 & 0.25 & 0.28 & 0.27 & 0.263 \\
LLaMA-2-7B-Chat & 0.16 & 0.16 & 0.14 & 0.17 & 0.18 & 0.20 & 0.168 \\ 
AquilaChat2-7B & 0.13 & 0.15 & 0.15 & 0.07 & 0.11 & 0.09 & 0.117 \\
\bottomrule
\end{tabular}
}
\caption{Performance on the FollowEval benchmark by the evaluated LLMs and human evaluators. ACC-$i$ represents the accuracy of the $i$-th run on the FollowEval benchmark, while AVG denotes the mean accuracy derived from three separate runs.}\label{test_results}
\end{table*}

\section{Experiments}
To validate the efficacy of the FollowEval benchmark in evaluating the instruction-following capabilities of LLMs, we have conducted comprehensive experiments utilizing the FollowEval benchmark. In this section, we offer a comprehensive overview of the experimental setup. This is followed by a thorough presentation and analysis of the experimental results, providing a deep understanding of the LLMs' performance in following instructions.

\subsection{Experimental Settings}

\subsubsection{Evaluated Models}
We evaluate a variety of representative LLMs on the FollowEval benchmark. These LLMs encompass both open-source and proprietary models, and have demonstrated remarkable performance across diverse benchmarks. A brief overview of the evaluated LLMs is provided below:
\begin{itemize}
    \item \textbf{GPT-4} \citep{DBLP:journals/corr/abs-2303-08774} represents the most advanced LLM in the GPT series. This model has been fine-tuned through reinforcement learning from human feedback (RLHF) \citep{DBLP:conf/nips/StiennonO0ZLVRA20,DBLP:conf/nips/Ouyang0JAWMZASR22} to generate responses that align with human preference and values. The superior performance of GPT-4 is demonstrated by its state-of-the-art performance across a diverse range of tasks and benchmarks, thereby solidifying its position as the most proficient general LLM currently available.
    \item \textbf{GPT-3.5-Turbo}\footnote{\url{https://platform.openai.com/docs/models/gpt-3-5}} is the most capable variant in the GPT-3.5 models and has been specifically optimized for conversational applications. This model has also been fine-tuned through RLHF, enabling it to follow to a variety of user instructions and generate detailed responses.
    \item \textbf{ChatGLM-6B series} represents the evolution of the open-source, bilingual (Chinese-English) conversational models with approximately 6 billion parameters. Our experiments involve both the second and third iterations of this series, namely ChatGLM2-6B\footnote{\url{https://huggingface.co/THUDM/chatglm2-6b}} and ChatGLM3-6B.\footnote{\url{https://huggingface.co/THUDM/chatglm3-6b}} These iterations incorporate several key enhancements over their predecessor, such as stronger performance, longer context, and more open license.
    \item \textbf{Qwen-Chat series} \citep{DBLP:journals/corr/abs-2309-16609} comprises two open-source models, Qwen-7B-Chat\footnote{\url{https://huggingface.co/Qwen/Qwen-7B-Chat}} and Qwen-14B-Chat\footnote{\url{https://huggingface.co/Qwen/Qwen-14B-Chat}}, with parameter counts of 7 billion and 14 billion, respectively. These models have been pretrained on up to 3 trillion tokens of multilingual data, with a particular emphasis on Chinese and English. Following pretraining, an alignment technique was employed during fine-tuning to ensure that the responses generated by these models align with human preferences.
    \item \textbf{Baichuan-Chat series} consists of three open-source variants: Baichuan-13B-Chat\footnote{\url{https://huggingface.co/baichuan-inc/Baichuan-13B-Chat}}, Baichuan-2-7B-Chat\footnote{\url{https://huggingface.co/baichuan-inc/Baichuan2-7B-Chat}}, and Baichuan-2-13B-Chat\footnote{\url{https://huggingface.co/baichuan-inc/Baichuan2-13B-Chat}}, containing 13 billion, 7 billion, and 13 billion parameters, respectively. The Baichuan-13B-Chat model is pretrained on a corpus of 1.4 trillion tokens, while the Baichuan-2-7B-Chat and Baichuan-2-13B-Chat models are pretrained on a more extensive corpus \citep{DBLP:journals/corr/abs-2309-10305}, comprising 2.6 trillion tokens. Similar to the aforementioned LLMs the models in the Baichuan-Chat series have been finetuned for human alignment.
    \item \textbf{InternLM-7B-Chat}\footnote{\url{https://huggingface.co/internlm/internlm-7b}} model, tailored for practical applications, possesses 7 billion parameters. It has undergone pretraining on trillions of tokens. Additionally, it supports a maximum context window size of 8k, which facilitates longer input and stronger reasoning capacities.
    \item \textbf{LLaMA-2-Chat series} constitutes a collection of open-source models with parameter counts ranging from 7 billion to 70 billion \citep{DBLP:journals/corr/abs-2307-09288}. These models have been trained on 2 trillion tokens of data and subsequently fine-tuned through RLHF to align with human preferences. We evaluate two models from this series: LLaMA-2-7B-Chat\footnote{\url{https://huggingface.co/meta-llama/Llama-2-7b-chat-hf}} with 7 billion parameters and LLaMA-2-13B-Chat\footnote{\url{https://huggingface.co/meta-llama/Llama-2-13b-chat-hf}} with 13 billion parameters, utilizing the FollowEval benchmark.
    \item \textbf{AquilaChat2-7B}\footnote{\url{https://huggingface.co/BAAI/AquilaChat2-7B}} is an open-source chat model from the AquilaChat2 series that encompasses 7 billion parameters. It has been trained through instruction tuning to enhance its interaction with humans.
\end{itemize}

Given that nucleus sampling \citep{DBLP:conf/iclr/HoltzmanBDFC20} and top-k sampling \citep{DBLP:conf/acl/LewisDF18,DBLP:conf/acl/ChoiBGHBF18,radford2019language}, which sample tokens from a truncated probability distribution at each timestep, are commonly adopted decoding strategies for LLMs, the LLMs can generate different responses even for identical input. Therefore, to mitigate the randomness in the LLMs' responses and enhance the reliability of the experimental results, each model is evaluated three times on the FollowEval benchmark.

In addition to the LLMs we evaluate, to establish a strong human baseline, we also recruit three individuals with diverse cultural and educational backgrounds and use the FollowEval benchmark to test these individuals in a simulated examination.

\subsubsection{Evaluation Metrics}
\begin{CJK*}{UTF8}{gbsn}
For the FollowEval benchmark, accuracy is employed as the evaluation metric. This metric quantifies the proportion of test instances that are correctly responded to by the LLMs. To determine the correctness of the responses, we manually design rules for each test example.
For instance, in a test instance with strict response constraints, such as ``若a > b，b > c，请问a > c吗，直接回答正确还是错误，不要输出分析过程和额外的其他字符\ (\textit{If `a' is greater than `b', and `b' is greater than `c', then is it correct to say that `a' is greater than `c'? Please respond with `correct' or `incorrect' directly, without including any analytical process or extra characters})'', the LLMs should generate either “正确\ (\textit{correct})” or “错误\ (\textit{incorrect})”, excluding any other text. If the LLMs' responses do not match ``正确\ (\textit{correct})'' or ``错误\ (\textit{incorrect})'', they are deemed incorrect.
On the other hand, in a test instance with less rigorous response constraints, such as ``小芳喜欢李华、王刚，不过她更喜欢赵明，请问小芳最喜欢谁?\ (\textit{Xiaofang likes both Li Hua and Wang Gang, but she prefers Zhao Ming. So, who is Xiaofang's favorite?})'', the LLMs' responses are considered correct if they include ``李华\ (\textit{Li Hua})'', ``王刚\ (\textit{Wang Gang})'', or ``赵明\ (\textit{Zhao Ming})''.
\end{CJK*}

\subsection{Experimental Results}
The experimental results of various LLMs on the FollowEval benchmark are presented in Table~\ref{test_results}. These results reveal that humans achieve a flawless accuracy rate of 100\% on the FollowEval benchmark, underscoring their impeccable ability to adhere to the instructions of this benchmark. In comparison, GPT-4, despite being the top-performing model among the large LLMs we tested, falls short of matching human performance. Furthermore, the proprietary models evaluated (GPT-4 and GPT-3.5-Turbo) exhibited superior instruction-following ability compared to their open-source counterparts. Moreover, a clear trend emerges within the same series of LLMs: those with a higher parameter count typically outperform their counterparts with fewer parameters on the FollowEval benchmark. This suggests a positive correlation between the size of the model and its performance on the FollowEval benchmark.

\section{Conclusion}

In this paper, we introduce FollowEval, a new benchmark for evaluating the instruction-following capabilities of LLMs in both English and Chinese. FollowEval comprises 200 manually curated test instances designed to assess LLMs across diverse dimensions of instruction-following, including string manipulation, commonsense reasoning, logical reasoning, spatial reasoning, and adherence to response constraints. Each test instance incorporates multiple essential elements that must be adequately addressed for a fully correct response. To facilitate automatic evaluation, handcrafted rules are provided for each example. Comprehensive experiments demonstrate a substantial gap between LLMs and human performance on FollowEval, underscoring areas requiring further advancement.

\clearpage
\bibliography{anthology}

\appendix



\end{document}